\documentclass[sigconf]{acmart}
\usepackage{algorithm}
\usepackage{algorithmic}
\usepackage{newfloat}
\usepackage{listings}
\usepackage{tabularx}
\usepackage{amsmath}
\usepackage{booktabs}
\usepackage{multirow}
\usepackage{amsthm}
\usepackage{subfig}
\usepackage{xcolor}
\usepackage{arydshln}
\usepackage{enumitem}
\usepackage{arydshln}
\usepackage{stfloats}

\newtheorem{definition}{Definition}
\newtheorem{problem}{Problem}
\newtheorem{theorem}{Theorem}
\newtheorem{lemma}{Lemma}

\newcommand{\YHY}[1]{{\color{black}{#1}}}

\newcommand{\hide}[1]{}
\newcommand{\vpara}[1]{\vspace{0.05in}\noindent\textbf{#1 }}

\AtBeginDocument{%
  \providecommand\BibTeX{{%
    \normalfont B\kern-0.5em{\scshape i\kern-0.25em b}\kern-0.8em\TeX}}}

\copyrightyear{2024}
\acmYear{2024}
\setcopyright{acmlicensed}
\acmConference[KDD '24] {Proceedings of the 30th ACM SIGKDD Conference on Knowledge Discovery and Data Mining }{August 25--29, 2024}{Barcelona, Spain}
\acmBooktitle{Proceedings of the 30th ACM SIGKDD Conference on Knowledge Discovery and Data Mining (KDD '24), August 25--29, 2024, Barcelona, Spain}
\acmDOI{10.1145/3637528.3672013}
\acmISBN{979-8-4007-0490-1/24/08}

\begin{document}

\title{Unveiling Privacy Vulnerabilities: \\Investigating the Role of Structure in Graph Data}


\author{Hanyang Yuan}
\authornote{State Key Laboratory of Blockchain and Security. The author is also at Hangzhou High-Tech Zone (Binjiang) Institute of Blockchain and Data Security, Hangzhou, China. This work was done when the author was a visiting student at Fudan University.}
\affiliation{%
  \institution{Zhejiang University}
  \city{Hangzhou}
  \country{China}}
\affiliation{%
  \institution{Fudan University}
  \city{Shanghai}
  \country{China}}
\email{yuanhanyang@zju.edu.cn}

\author{Jiarong Xu}
\authornote{Corresponding author.}
\affiliation{%
  \institution{Fudan University}
  \city{Shanghai}
  \country{China}}
\email{jiarongxu@fudan.edu.cn}

\author{Cong Wang}
\affiliation{
 \institution{Peking University}
 \city{Beijing}
 \country{China}}
\email{wangcong@gsm.pku.edu.cn}

\author{Ziqi Yang}
\affiliation{%
  \institution{Zhejiang University}
  \city{Hangzhou}
  \country{China}}
\email{yangziqi@zju.edu.cn}

\author{Chunping Wang}
\affiliation{%
  \institution{Finvolution Group}
  \city{Shanghai}
  \country{China}
}
\email{wangchunping02@xinye.com}

\author{Keting Yin}
\affiliation{%
  \institution{Zhejiang University}
  \city{Hangzhou}
  \country{China}}
  \email{yinkt@zju.edu.cn}

\author{Yang Yang}
\affiliation{%
  \institution{Zhejiang University}
  \city{Hangzhou}
  \country{China}}
\email{yangya@zju.edu.cn}

\renewcommand{\shortauthors}{Hanyang Yuan et al.}




\begin{abstract}
The public sharing of user information opens the door for adversaries to infer private data, leading to privacy breaches and facilitating malicious activities. While numerous studies have concentrated on privacy leakage via public user attributes, the threats associated with the exposure of user relationships, particularly through network structure, are often neglected. 
This study aims to fill this critical gap by advancing the understanding and protection against privacy risks emanating from network structure, moving beyond direct connections with neighbors to include the broader implications of indirect network structural patterns.
To achieve this, we first investigate the problem of Graph Privacy Leakage via Structure (GPS), and introduce a novel measure, the Generalized Homophily Ratio, to quantify the various mechanisms contributing to privacy breach risks in GPS. 
Based on this insight, we develop a novel graph private attribute inference attack, which acts as a pivotal tool for evaluating the potential for privacy leakage through network structures under worst-case scenarios.
To protect users' private data from such vulnerabilities, we propose a graph data publishing method incorporating a learnable graph sampling technique, effectively transforming the original graph into a privacy-preserving version. 
Extensive experiments demonstrate that our attack model poses a significant threat to user privacy, and our graph data publishing method successfully achieves the optimal privacy-utility trade-off compared to baselines.

\end{abstract}
\maketitle
\begin{figure}[t]
    \centering
    {\includegraphics[width=1.0\columnwidth]{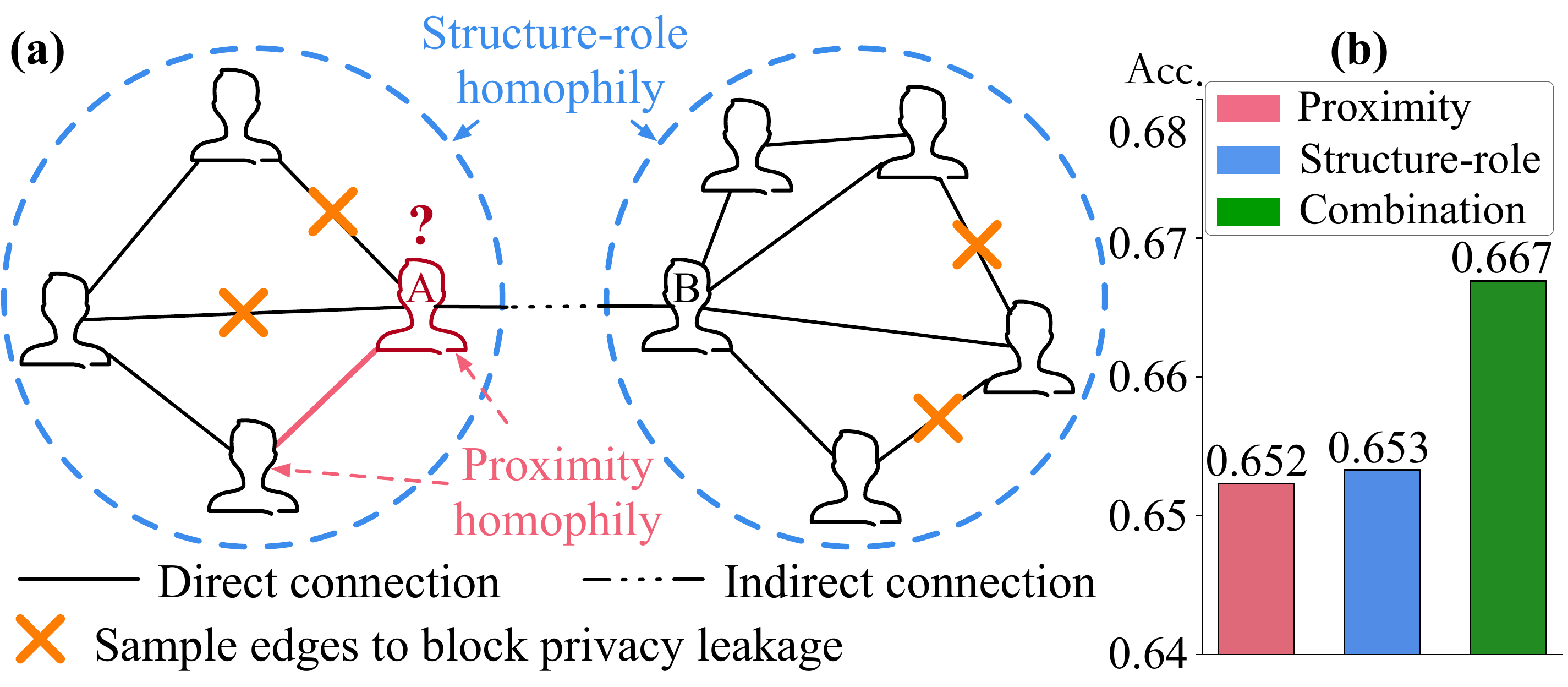}}
    \caption{(a) Illustration of privacy leakage mechanisms: proximity homophily highlighted in pink, structure-role homophily in blue, alongside the privacy protection strategy depicted in orange. 
    (b)  The results of private attribute inference attacks accounting for proximity homophily, structure-role homophily, and a combination of both on Pokec-n.}
    \label{fig: intro}
    \vspace{0.2in}
\end{figure}

\section{Introduction}

In the era of big data, with the increasing involvement of personal data in information technology and shared on the web, privacy protection has emerged as a crucial concern \cite{jia2018attriguard, ke2023privacy, ogbuke2022big}. In real-world scenarios, individuals often share some information publicly while safeguarding their private attributes. 
However, publicly available user information gives adversaries opportunities to infer private attributes,  resulting in privacy breaches~\cite{otterbacher2010inferring,rao2010classifying, mislove2010you}.
Furthermore, the inferred private data can facilitate malicious activities. For instance, in the 2010s, Cambridge Analytica collected personal data from 87 million Facebook users to infer their political stands, which were further used for political advertising, resulting in a scandal with over \$100 billion in economic losses~\cite{isaak2018user}. 
This incident emphasizes the urgent need for privacy protection mechanisms.

To prevent the exposure of user privacy, traditional works typically focus on privacy leakage through users' public attributes~\cite{jia2018attriguard,shokri2016privacy,salamatian2015managing}.
However, these methods often overlook the privacy risks stemming from the public relationships among users~\cite{jia2017attriinfer, mislove2010you, zhang2022mli}. 
{For example, in an online social platform like Facebook, users often publicly display their followers or friends, where these relationships collectively form a network.} This network structure can also give rise to potential privacy leakage \cite{zhang2022mli,mislove2010you, jia2017attriinfer}.

This work delves into the problem of Graph Privacy leakage via Structure (GPS), aimed at unveiling various mechanisms by which network structure can lead to privacy exposure.
Extant works on privacy breach through network structure are primarily premised in social homophily theory~\cite{mcpherson2001birds} that posits users with similar private attributes tend to connect with each other. Hence, they study privacy leaks through direct connections with neighbors~\cite{zhang2022mli, mislove2010you, jia2017attriinfer, hsieh2021netfense}.
\YHY{However, user privacy in networks can also be compromised through more complex structural patterns, extending beyond direct neighbors.}
Figure~\ref{fig: intro} (a) illustrates that privacy risks for user ``A'' arise not only from direct neighbors (\emph{i.e.}, proximity information), but also from users like ``B'' who, despite not being direct connections, exhibit similar local structures (\emph{i.e.}, structure-role information).
\YHY{An example of structure-role information in social networks is the observation that younger and older users  tend to maintain social circles of different sizes \cite{dong2014inferring}.}
This dimension of privacy leakage, facilitated by such local structures, has not been thoroughly investigated in existing studies. 
To fill this research gap, we aim to develop \emph{a graph data publishing method aimed at comprehensively protecting against potential privacy breaches arising from the network structure.} 
Nevertheless, achieving this goal presents several challenges.

The first challenge lies in how to measure the extent of privacy exposure through network structures.
Previous research mainly focus on privacy leakage through direct connections between nodes, using homophily to quantify proximity-related exposure
\cite{jiang2022topology,liu2022ud,suresh2021breaking}. However, this approach falls short in assessing privacy risks from structure-role information. 
In this study, we introduce the Generalized Homophily Ratio (GHRatio), a novel measure to quantify privacy risks associated with network structures.
The GHRatio is a general form that is adaptable to various structural features. We explore two prevalent cases---proximity homophily, structure-role homophily and their combination--- that contribute to privacy leakage.

The second challenge stems from the necessity to develop a private attribute inference attack model that utilizes proximity homophily, structure-role homophily, and their combination to launch attacks. Given that existing attack strategies merely exploit proximity homophily~\cite{al2012homophily, jia2017attriinfer, mislove2010you}, suboptimal results are yielded (as depicted by the red bar in Figure~\ref{fig: intro} (b)). 
To overcome this challenge, our model is designed to account for all identified privacy breaches through a data-centric strategy. This strategy involves providing a graph neural network (GNN) with various data forms, thus enhancing its capacity to learn from different types of homophily.
Consequently,  our attack model effectively behaves like a worst-case adversary, as evidenced by the green bar in Figure~\ref{fig: intro} (b).

The last challenge lies in how to design a graph data publishing approach that can effectively defend the worst-case private attribute inference attack.
Previous efforts in graph data publishing have primarily focused on differential privacy (DP) \cite{zhou2022structure, ye2020lf} and graph sampling \cite{hsieh2021netfense, cai2016collective}, but DP often compromises the utility of the data~\cite{li2009tradeoff}. In addition, many sampling methods are rule-based and reliant on domain-specific knowledge \cite{majeed2022comprehensive, majeed2020anonymization}, which restricts their applicability.
We therefore propose a learnable graph sampling method for privacy protection, employing a generative network that selectively samples edges to block privacy leakage (as illustrated by  $\vcenter{\hbox{\includegraphics[width=1.8ex,height=1.8ex]{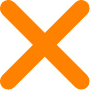}}}$ in Figure~\ref{fig: intro} (a)). This method ultimately produces a sampled graph suitable for publication.

Our contributions are summarized as follows:
\begin{itemize}[leftmargin=*]
    \item \textbf{Problem and measure}: Our work pioneers a comprehensive investigation into the problem of Graph Privacy leakage via Structure (GPS), introducing the innovative Generalized Homophily Ratio (GHRatio) as a measure of privacy leakage. 
    This helps us unveil all identified mechanisms by which the network structure can lead to privacy breaches in a quantitative manner. 
    \item  \textbf{Attack model}: We introduce a novel private attribute inference attack leveraging a data-centric strategy to exploit all identified privacy breaches. By feeding a GNN various data forms, it gains the ability to learn from multiple homophily types that result in privacy risks. 
    \item \textbf{Defensive model}:  To counter the attacks, we propose a graph data publishing method that employs learnable graph sampling, rendering the sampled graph suitable for publication.
    \item \textbf{Extensive experiments}:  Experiments in five real-world scenarios
    demonstrate {that (1) our private attribute inference attack beats the best baselines by an average of {+2.93\%}}, and (2) our graph data publishing approach achieves
    the optimal privacy-utility trade-off, \YHY{outperforming existing defensive methods when evaluated against worst-case attack scenarios.}
\end{itemize}
\vspace{0.15in}

\section{Problem Definition}\label{sec:problem}

Network structure is a significant factor contributing to privacy breaches, which stands as a fundamental data source for various network analysis tasks, such as node classification and community detection \cite{xu2021netrl, xu2020robust}.
Previous research has proven that modifying network structure is more effective than modifying node attributes in enhancing privacy protection \cite{hsieh2021netfense}. Therefore, this work particularly focuses on a privacy-preserving graph data publishing problem with a specific emphasis on the network structure.

In this study, we refer to privacy as a particular attribute that nodes choose to keep hidden, which aligns with previous works~\cite{li2020adversarial, hsieh2021netfense}. 
Let graph $G=(V, E, X)$ denote an undirected network, where $V=\{v_1,...,v_n\}$ is the node set, $E \subseteq V \times V$ is the edge set, and $X \in \mathbb{R}^{n \times m}$ is the node attribute matrix. $A \in \mathbb{R}^{n\times n}$ is the adjacency matrix of $G$, where $A_{ij} = 1$ if there exists an edge $(i,j) \in E$, otherwise $A_{ij}=0$. Each node $v_i \in V$ is associated with a known/unknown private attribute $Z_i$. Here,  $Z_i \in Z = Z_L \cup  Z_U$, where $Z_L$ denotes publicly available private attributes, and $Z_U$ represents hidden private attributes. 
In this context, privacy in the graph $G$ is defined by the set of hidden private attributes $Z_U$.

\YHY{
Let us consider the following attack scenario.
In a public social network, some users choose to conceal their private attributes, while others make them public. The adversary aims to infer these hidden private attributes.
The adversary is assumed to have access to the network structure, node attributes (typically non-private), and publicly available private attributes.
Publicly available private attributes can come from users who do not consider this information private or who seek to maximize visibility by sharing extensive personal information.}

Formally, we define this as graph private attribute inference attack problem.
\begin{problem}
[\textbf{Graph Private Attribute Inference Attack}]
Given graph $G=(V, E, X)$ and the publicly available private attributes $Z_L$, the graph private attribute inference attack aims to learn a function
\begin{equation}
f:G,Z_L \rightarrow Z_U,
\end{equation}
that predicts the hidden private attribute $Z_U$.
\end{problem}

The primary objective of this work is to tackle the problem of privacy-preserving graph data publishing against the aforementioned graph private attribute inference attack. Specifically,
instead of directly releasing the original graph, the data publisher is encouraged to generate a sampled graph for publishing, such that the sampled graph can defend against graph private attribute inference attack. Based on the above definition, we formulate our privacy-preserving graph data publishing problem as follows.

\begin{problem}
[{\textbf{Privacy-preserving graph data publishing}}] \label{def: defender goal}
Given graph $G=(V, E, X)$ and the publicly available private attributes $Z_L$, the data publisher aims to sample a new graph $G^\prime =(V, E^\prime, X)$ by selectively removing edges in $E$, resulting in a new edge set $E^\prime$. The sampled graph $ G^\prime $ is expected to simultaneously achieve the following two objectives:
 
\textbf{Objective 1: privacy preservation.} The adversary with $G^\prime$ and $Z_L$ cannot accurately infer the private attribute $Z_U$, \emph{i.e.},
\begin{equation}
\min_{G^\prime} \quad \mathrm{perf} \left(f (G^\prime), Z_U \right),
\end{equation}
where $ \mathrm{perf}  $ denotes a performance metric to evaluate how well the predicted value $f (G^\prime, Z_L)$ aligns with the ground truth $ Z_U $, such as accuracy or ROC-AUC as used in our work.

\textbf{Objective 2: utility.} 
The sampled graph $G^\prime$ should not deviate too much from the original graph $G$. This ensures that $G^\prime$ conveys useful information.
\end{problem}
\vspace{0.15in}

\section{Graph Privacy-leakage via structure (GPS)}\label{sec:homophily}

In this section, we delve into the problem of GPS, examining how the structure of a graph can potentially lead to privacy breaches. 
We aim to introduce a novel measure to quantify the various mechanisms contributing to privacy breach risks in GPS.

\begin{figure}[t]
    \centering
    {\includegraphics[width=1.0\columnwidth]{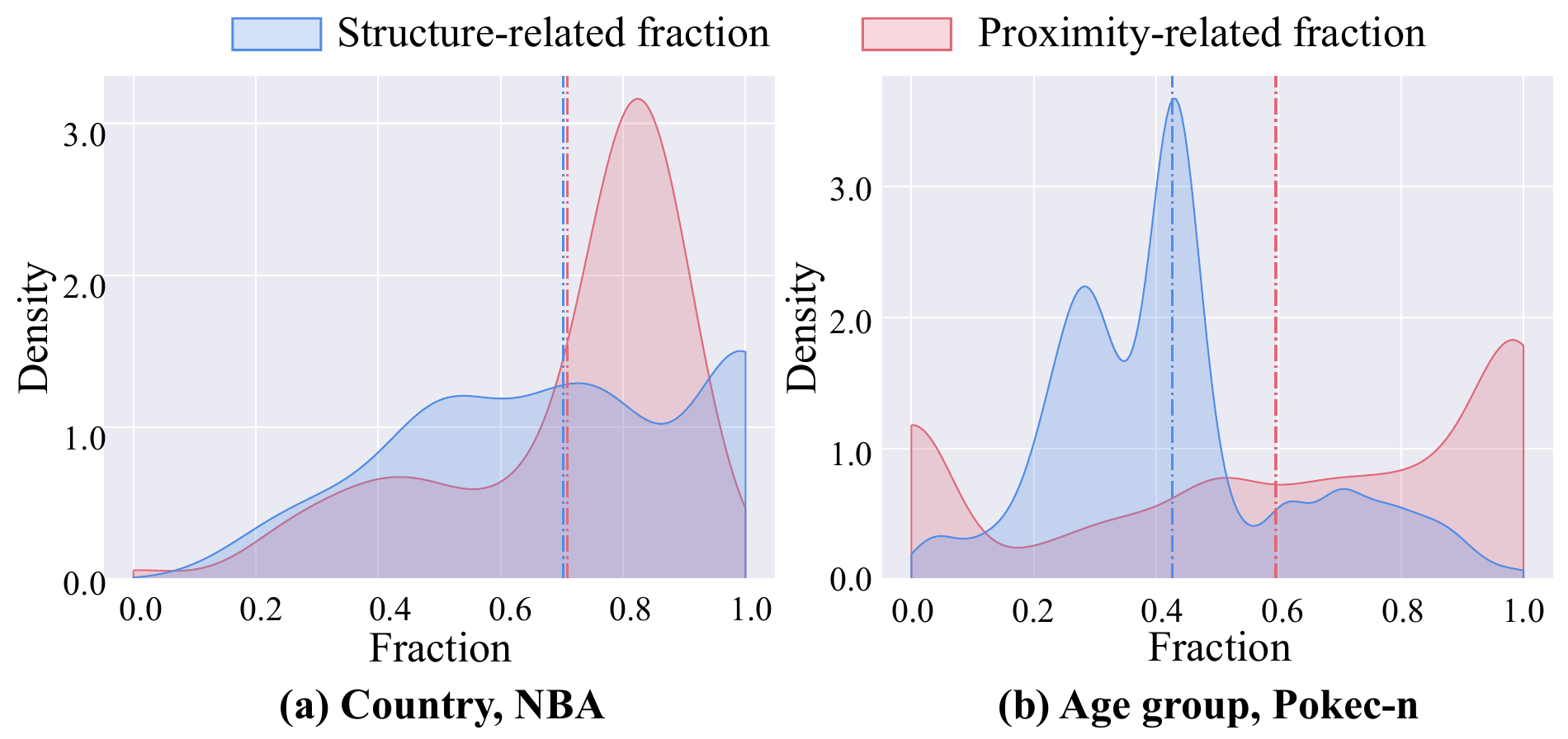}}
    \caption{Visualization of proximity-related fraction and structure-related fraction distributions on NBA
    and Pokec-n.
    }
    \label{fig: observation}
    \vspace{0.2in}
\end{figure}

We start with an exploratory analysis using the NBA and Pokec-n datasets, two widely adopted datasets for graph privacy-preserving learning \cite{dai2021say, ling2022learning, dai2022learning}. We calculate (1) proximity-related fraction: the fraction of a node's neighbors sharing the same private attribute; and (2) structure-related fraction: the fraction of nodes with similar local structures and the same private attributes to the total nodes with similar local structures for a specific node.
\YHY{Here, similar local structures are defined where the structural similarity between nodes' ego networks exceeds a set threshold, assessed using degree centrality.}
From the results shown in Figure~\ref{fig: observation}, we derive two key observations:

\emph{Observation 1:  The necessity for node-level analysis.}
In Figure~\ref{fig: observation} (a), despite similar mean values for both fractions, their local distributions differ. 
\YHY{Figure~\ref{fig: observation} (b) shows that even when the mean of the proximity-related fraction dominates, some nodes have significantly lower values of the proximity-related fraction. }
This challenges the graph-level homophily ratios from prior studies, which rely on mean values as indicators~\cite{zhu2020beyond, lim2021large}, underscoring the need for node-level analysis.

\emph{Observation 2: Proximity, structure, and their combination should be simultaneously considered.}
We find that in Figure~\ref{fig: observation} (a), there are nodes in which both fractions are relatively high; 
\YHY{and in Figure~\ref{fig: observation} (b), there are instances where the proximity-related fractions of certain nodes are small, while the structure-related fraction may provide supplementary information.} These observations emphasize the importance of simultaneously considering proximity, structure, and their combination when addressing privacy concerns. 

In light of the insights gained from the exploratory analysis, we propose
a novel measure known as the Generalized Homophily Ratio (GHRatio), which is a generalized form that can be used in conjunction with different definitions of structural features associated with graph privacy.
Subsequently, we instantiate three forms of GHRatio: proximity homophily, structure-role homophily, and their combination. They represent the main pathways in graph structure through which privacy can be leaked.

\vpara{Generalized Homophily Ratio.}
As our goal is to investigate how network structure discloses privacy, we begin by defining homophily indicator, which can be used to characterize the structural characteristic or relation between two nodes.

\begin{definition}[Homophily indicator] \label{def: homophily indicator}
A homophily indicator $\delta(v_i,v_j,r)$ assesses whether node $v_i$ and $v_j$  exhibit a shared structural characteristic or relation $r$. For example, 
when $r$ signifies  similar local structure, we have $\delta(v_i,v_j,r)=1$ if the similarity of $v_i$ and $v_j$ exceeds a certain threshold, and 0 otherwise;
when considering $r$  as the relation of adjacency, $\delta(v_i,v_j,r)=1$ if node  $v_i$  and $v_j$ are directly connected, and 0 otherwise.
\end{definition}
Based on the homophily indicator, we can define the GHRatio as follows.

\begin{definition}[Generalized Homophily Ratio (GHRatio)] \label{def: GHRatio}
Given an observed graph $G$ and private attribute matrix $Z$, the GHRatio of node $v_i$ is defined as  the conditional probability that node $v_i$ and any node $v_j$ ($j \neq i$) have the same private attributes, given that $\delta(v_i,v_j,r)=1$, \emph{i.e.},
\begin{equation} \label{eq: GHRatio}
    \mathrm{GHRatio}_i =  P(Z_i = Z_j | \delta(v_i,v_j,r)=1), j \neq i,
\end{equation}
where $Z_i$ and  $Z_j$ are private attributes of  node $v_i$ and $v_j$, respectively.
\end{definition}

\vpara{Two prevalent cases of GHRatio.}  
Given the general form of GHRatio, we further delve into two prevalent cases of it: proximity homophily, structure-role homophily, by defining specific graph structural feature associated with GHRatio. 

\textbf{(1) Proximity homophily ratio.} 
By defining  the $r$ in 
homophily indicator $\delta(v_i,v_j,r)$  as
the relation of adjacency, we have $\delta(v_i,v_j,r)=1$ if $v_i$ and $v_j$ are connected, and 0 otherwise;
We name this specific case of GHRatio as Proximity Homophily Ratio ($\mathrm{GHRatio}^\text{prox}$):

\begin{equation} \label{eq: feature homo}
\mathrm{GHRatio}_i^\text{prox}
=\frac{|\{j|j \in \mathcal{N}(i) \land Z_j = Z_i\} |}{|\{j|j \in \mathcal{N}(i) \}|},
\end{equation}
where $\mathcal{N}(i)$ denotes $v_i$'s neighborhood, and $|\cdot|$ denotes the cardinality of a set. In fact, $\mathrm{GHRatio}^\text{prox}$ aligns with the node-level homophily ratio defined in existing works \cite{liu2022ud, suresh2021breaking}.

\textbf{(2) Structure-role homophily ratio.} {We define $r$ in the homophily indicator $\delta(v_i, v_j, r)$ as the similar local structure. Then, we have $\delta(v_i, v_j, r) = 1$ if the similarity between the local structures of $v_i$ and $v_j$ exceeds a certain threshold, and 0 otherwise. We name this case of GHRatio as Structure-Role Homophily Ratio ($\mathrm{GHRatio}^\text{role})$:

\begin{equation} \label{eq: structure homo}
\mathrm{GHRatio}_i^\text{role}=\frac{| \{j|g(j) \sim g(i) \land Z_j = Z_i\}| }{|\{j|g(j) \sim g(i) \}|},
\end{equation}
where $g(i)$ is the ego network of node $v_i$, $g(i) \sim g(j)$ denotes that the ego networks  $g(i)$ and  $g(j)$ are sufficiently similar (\emph{e.g.}, this similarity can be understood in terms of structural similarity, specifically when the structural similarity between  $g(i)$ and  $g(j)$ exceeds a predetermined threshold).
Empirically, we adopt the degree centrality to characterize the similarity between ego networks, which have been validated for its strong effectiveness and efficiency in previous studies \cite{ying2021transformers, ribeiro2017struc2vec}.}
\vspace{0.15in}

\section{Private Attribute Inference Attack}\label{sec:attack}

This section introduces a novel private attribute inference attack model that leverages proximity homophily, structure-role homophily and their combination to disclose private information.
We adopt a data-centric approach, feeding varied data forms into a GNN to extract the representations related to different homophily types (see \S~\ref{subsec:data}).
Then, a routing operator is introduced for the adaptive integration of these homophily-related representations (see \S~\ref{subsec:routing}). 
See Figure~\ref{fig: model} (a) for an overview of our proposed attribute inference attack.

\subsection{Enhancing GNNs with Data-Centric Strategy}\label{subsec:data}

Developing a GNN model capable of learning representations tailored to different homophily types is challenging.
Existing works predominantly learn representations based on proximity information \cite{hamilton2017inductive, kipf2016semi} or high-order node dependencies \cite{abu2019mixhop, zhu2020beyond}, which can not adequately learn structure-role information.  Although some network representation methods \cite{ribeiro2017struc2vec, ahmed2018learning, henderson2012rolx} are designed to learn from node structure roles, their expressive power is limited.

\begin{figure}[t]
    \centering    {\includegraphics[width=0.99\columnwidth]{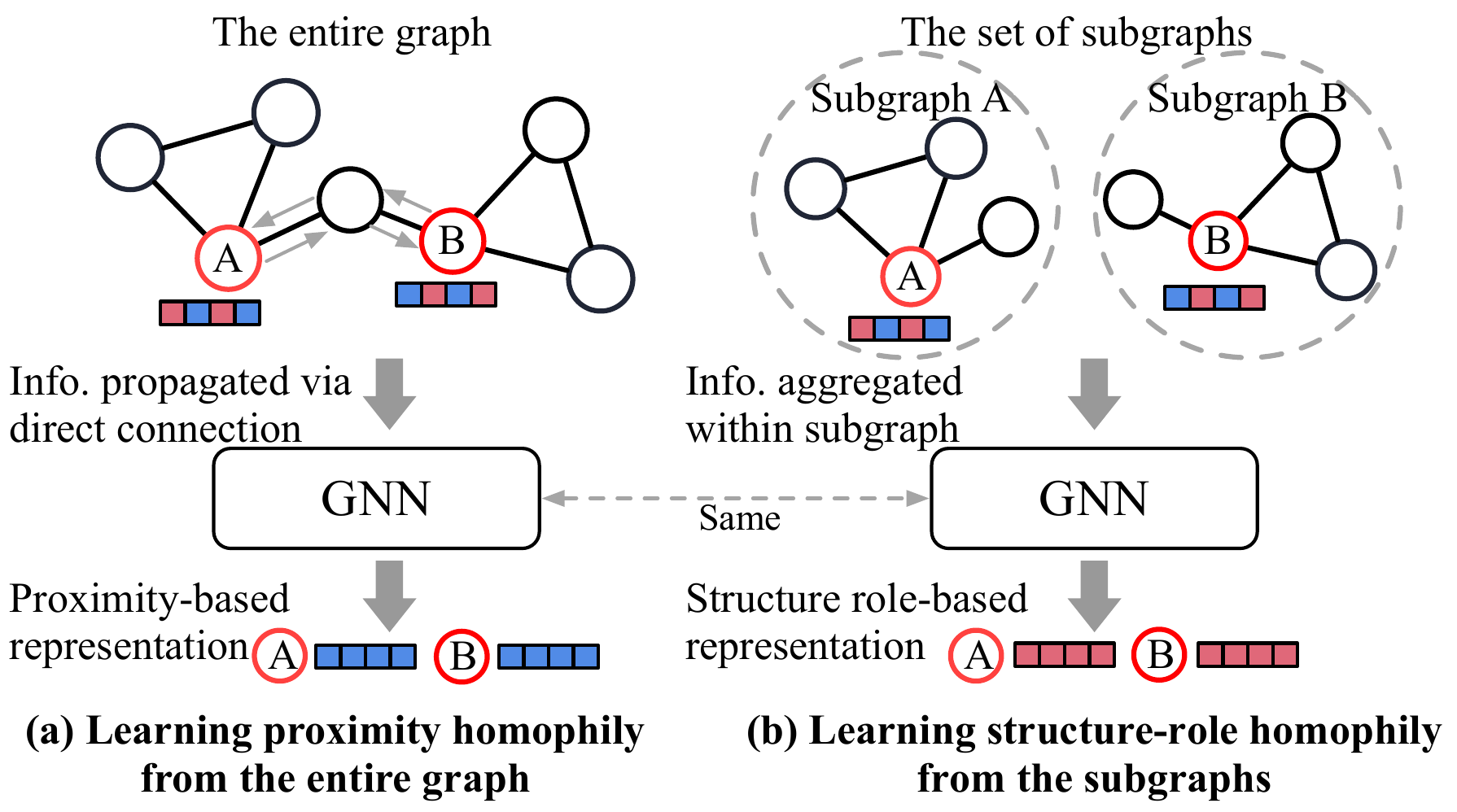}}
    \caption{Illustration of our data-centric strategy of
    feeding different data forms (\emph{i.e.}, graph vs subgraphs) into GNN to learn different knowledge  (\emph{i.e.}, proximity homophily vs structure-role homophily).
    }
    \label{fig: Illustration of our data-centric approach}
    \vspace{0.2in}
\end{figure}

In this work, we introduce a data-centric strategy designed to enhance GNNs' capacity to learn representations tailored to proximity homophily and structure-role homophily. 
This is achieved by feeding different forms of data into GNNs. An illustrative example is provided in Figure~\ref{fig: Illustration of our data-centric approach}. Our key insight is:
\begin{enumerate}[leftmargin = 10pt]
    \item Feeding the entire graph to a GNN enables it to learn proximity homophily;
    \item Feeding the set of nodes' subgraphs (\emph{e.g.}, ego networks) to a GNN, where each node's representation is computed as its subgraph's representation, facilitates the learning of structure-role homophily.
\end{enumerate} 

\vpara{Learning proximity homophily from the entire graph.}
This principle aligns with the message-passing mechanism in standard GNNs, where nodes update their representations by aggregating information from their direct neighbors~\cite{wu2020comprehensive, kipf2016semi}. This aggregation process naturally encourages similarity in the representations of connected nodes~\cite{wu2020comprehensive}.
Specifically, we generate proximity-based representations $H^\mathrm{prox} \in \mathbb{R}^{n \times d}$ by feeding the entire graph $G$ to a GNN encoder  $\mathrm{GNN^{prox}}$, \emph{i.e.},
\begin{equation}
    H^\mathrm{prox} = \mathrm{GNN^{prox}} \left(G \right).
\end{equation}

\vpara{Learning structure-role homophily from subgraphs.}
This approach first extracts $k$-hop ego networks for each node, resulting in a set of subgraphs. These subgraphs are subsequently fed to a GNN encoder.
The resulting subgraph representation, which encapsulates the aggregated features of its constituent nodes,  serves as the representation for the subgraph's central node.

This benefits the GNN's ability to learn structure-role homophily for two main reasons:
First, by concentrating on subgraphs, the central node's representation becomes exclusively reflective of its local structural context,  isolating the central node from external influences of nodes outside the subgraph. 
Secondly, GNNs are particularly adept at learning structural information of smaller subgraphs.  Although GNNs can identify structural patterns, comprehending complex structures in larger graphs remains challenging, as observed in \cite{huang2020graph}.
Ultimately, this process allows the GNN to bring the representations of nodes with similar local structures closer in the latent space.

Specifically, we process the $k$-hop ego network of node $i$, denoted as $S_i = (V_i, A_i, X_i)$, and feed the subgraph to a GNN encoder $\mathrm{GNN^{role}}$. The node representation $H^\mathrm{role}_i$ is computed as the subgraph representation, which is the mean of the representations of all nodes in the subgraph:
\begin{equation}
    H^\mathrm{role}_i = \mathrm{Pooling}(\mathrm{GNN^{role}}(S_i)),
\end{equation}
where $\mathrm{Pooling}$ is the mean pooling operator.

\vpara{Theoretical analysis of learning from subgraphs.}
While capturing proximity homophily from the entire graph is widely acknowledged \cite{wu2020comprehensive,khoshraftar2024survey,yang2022graph}, few works explore learning structure-role homophily from subgraphs. We therefore theoretically investigate whether GNNs fed with subgraphs can learn structure-role homophily.
The following theorem suggests that nodes with similar local structures can obtain similar node representations.
\begin{theorem}[] \label{thm: local structure similarity}
Let $S_{i}$ and $S_{j}$ be two k-hop subgraphs induced from node $v_i$ and $v_j$. After employing a $K$-layer GNN encoder with a 1-hop graph filter $\Psi(\mathcal{L})$ on each subgraph, the representations of the center node $v_i$ and $v_j$ are obtained via a pooling function, \emph{i.e.}, $H^\mathrm{role}_i = \mathrm{Pooling}(\mathrm{GNN^{role}}(S_i))$ and $H^\mathrm{role}_j = \mathrm{Pooling}(\mathrm{GNN^{role}}(S_j))$. Without loss of generality, assume that the attribute of each node is a vector of ones,  $H^\mathrm{role}_i$ and $H^\mathrm{role}_j$ satisfy:
$$
\| H^\mathrm{role}_i - H^\mathrm{role}_j\|_2 \leq \tau \|\mathcal{L}_{i}-\mathcal{L}_{j}\|_2,
$$
where $\| \cdot \|_2 $ denotes $L_2$ norm of matrix or vector, $\tau$ denotes a constant depending on $\mathrm{GNN^{role}}$, $\mathcal{L}_i$ denotes the normalised Laplacian matrix of $S_{i}$.
\end{theorem}
In Theorem \ref{thm: local structure similarity}, the term $\|\mathcal{L}_{i}-\mathcal{L}_{j}\|_2$ measures the difference of local structure around $v_i$ and $v_j$. As similar local structures bring smaller differences in Laplacian matrices, the upper bound of node representation distance is reduced. Consequently, nodes with similar local structures become closer in the latent space. The proof of this theorem can be found in Appendix \ref{sub: proof}.

\subsection{Routing Operator}\label{subsec:routing}
\begin{figure}[t]
    \centering
    {\includegraphics[width=1\columnwidth]{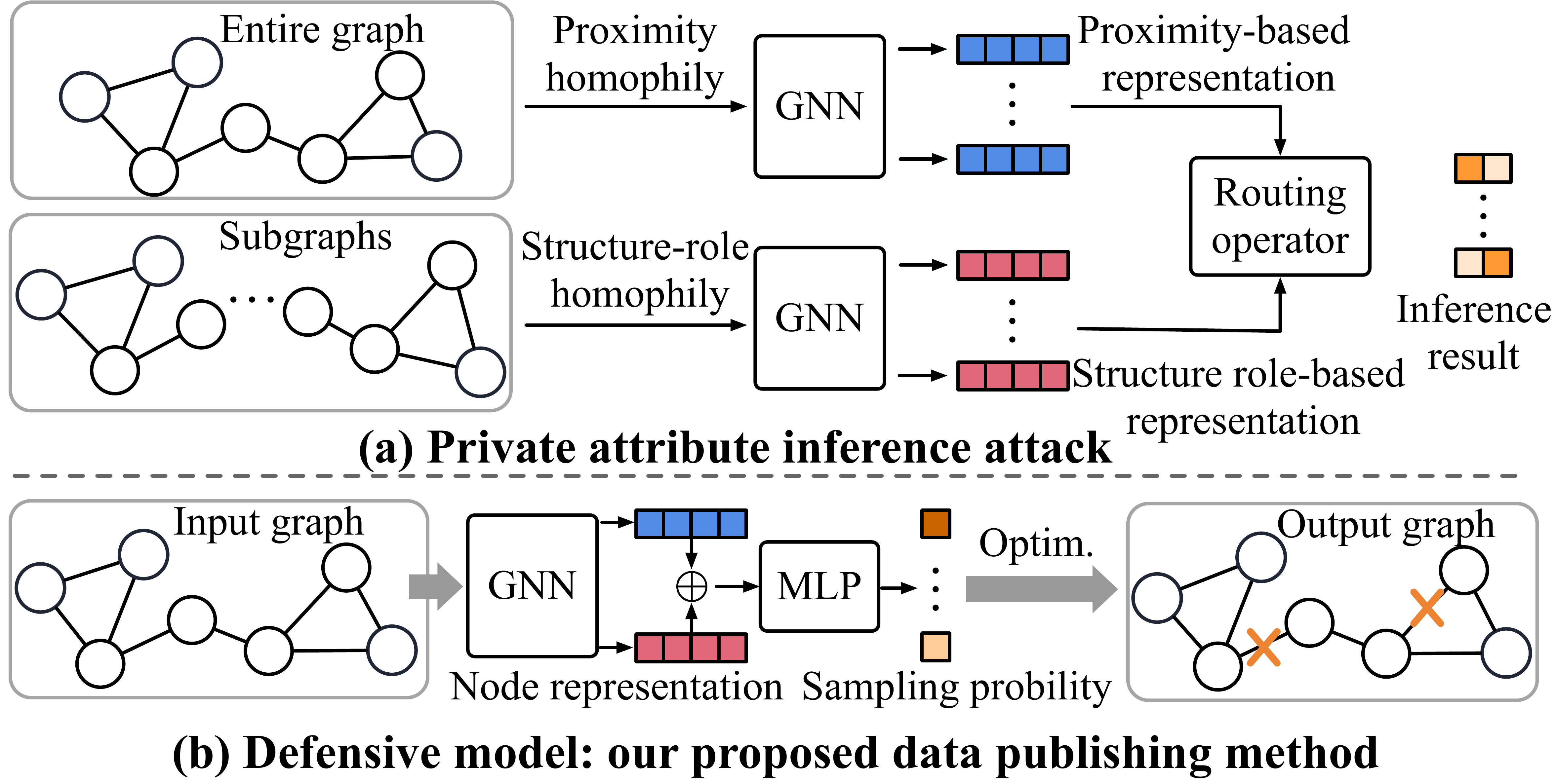}}
    \caption{An overview of (a) the proposed attribute inference attack and (b) the proposed graph data publishing method.}
    \label{fig: model}
    \vspace{0.2in}
\end{figure}
After obtaining the two types of representations, the challenge arises in determining the optimal method to integrate them, especially given the uncertainty about the extent of information that should be merged from each type. To tackle this issue, we introduce a \emph{routing operator} that leverages our proposed $\mathrm{GHRatio}$ to effectively combine these representations.

Since  $\mathrm{GHRatio}^\text{prox}$ and $\mathrm{GHRatio}^\text{role}$  serve to quantify the extent to which proximity and structural role disclose privacy, respectively. Utilizing these ratios, we can integrate the two types of representations, aiming for a balanced and informed combination that reflects the significance of both proximity and structural information in revealing privacy.
However, calculating these ratios requires the knowledge of the private attributes of all nodes, which presents another difficulty. To address this, our approach involves estimating the ratios by employing the pseudo-labels of private attributes.

Formally, we apply Multilayer Perceptrons (MLPs) to obtain inference results from proximity-based representation $H_\mathrm{prox}$ and structure role-based representation $H_\mathrm{role}$. Then we use the estimated GHRatios as the proportions to integrate them. The
integrated result $\hat{Z}_i$ in turn serves as the pseudo-labels.
Due to the interdependence between pseudo-labels and GHRatios, we initialize GHRatios with constants and iteratively update them during training.
The above process can be described as follows:
\begin{align}
&\hat{Z}_i^\mathrm{prox} = \mathrm{MLP} ( H^\mathrm{prox}_i), \hat{Z}_i^\mathrm{role} = \mathrm{MLP} ( H^\mathrm{role}_i), \notag \\ 
&\hat{Z}_i = \mathrm{GHRatio}_i^\text{prox} \cdot \hat{Z}_i^\mathrm{prox} +\mathrm{GHRatio}_i^\text{role} \cdot \hat{Z}_i^\mathrm{role}, 
\end{align}
where $\hat{Z}_i$ denotes the final inference result for node $v_i$. 

The model is optimized as follows:
\begin{equation}
\min_{\Theta} \quad L_{\mathrm{adv}}=
 - \frac{1}{|V_L|} \sum_{v_i \in V_L} \sum_{c=1}^C Z_{i,c} \log \hat{Z}_{i,c},
\end{equation}
where $C$ denotes the number of categories for the private attribute, $V_L$ denotes the nodes with known private attributes, and $\Theta$ denotes the parameters of the attack model.
\vspace{0.15in}

\section{Privacy-preserving Graph Data Publishing}\label{sec:defense}

In this section, we introduce a learnable graph sampling approach for privacy-preserving graph data publishing. We first outline the learnable strategy for graph sampling in \S~\ref{subsec: parameterization}. Following this, the optimization objectives and training algorithm are detailed in \S~\ref{subsec: optimization} and \S~\ref{subsec: algorithm}, respectively.
Figure~\ref{fig: model} (b) provides an overview of the proposed graph data publishing method.

\subsection{Learnable Graph Sampling} \label{subsec: parameterization}

To generate a sampled graph $G'$ with adjacency matrix $A'$,  we first feed the graph  $G$  into a GNN encoder to obtain node representation $H^\mathrm{samp}$: 
\begin{equation}
H^\mathrm{samp} = \text{GNN}(G).
\end{equation}
Given any connected node pair $\{v_i, v_j\}$, an MLP with sigmoid activation takes the concatenation of their node representations $H^\mathrm{samp}_i$ and $H^\mathrm{samp}_j$ as input to compute the probability $\mathcal{T}_{ij}$ of preserving the edge between $\{v_i, v_j\}$:
\begin{equation} \label{eq: sampling probability}
\mathcal{T}_{ij} = \sigma (\text{MLP} (H^\mathrm{samp}_i \oplus H^\mathrm{samp}_j)),
\end{equation}
where $\sigma$ denotes sigmoid activation, and $\oplus$ represent the concatenation operation.
With an edge's $\mathcal{T}_{ij}$ calculated, we sample whether to retain the edge in the synthetic graph $G'$ according to this probability, where
\begin{equation}
A^{\prime}_{ij} \sim \mathrm{Bernoulli}(\mathcal{T}_{ij}).
\end{equation}
In particular, the Gumbel-Softmax reparameterization trick \cite{maddison2016concrete} is utilized to tackle the non-differentiable nature of the sampling process. In doing so, we obtain a continuous sampling result, \emph{i.e.}, 
$A^{\prime}_{ij} =\sigma ( ( \log U-\log \left(1-U \right)+\log \mathcal{T}_{ij}) / \varepsilon )$,
where $U \sim \mathrm{Uniform}(0,1)$. As the temperature hyper-parameter $\varepsilon$ tends to zero, the reparameterized result smoothly converges to binary values, while maintaining the relative order of each Gumbel \cite{maddison2016concrete}.

\subsection{Optimization Problem} \label{subsec: optimization}
We propose three optimization objectives for training learnable parameters within the graph sampling procedure: one aimed at defending against worst-case attacks, one designed for a broader spectrum of attacks, and another dedicated to preserving essential graph properties. These objectives collectively ensure that the sampled synthetic graph maintains user privacy while simultaneously achieving desirable data utility.

\vpara{Defending Against Worst-Case Attack.}
Given the proposed inference attack, the most straightforward and effective approach is to defend against this attack under the worst-case. Specifically, to obtain the worst-case attack, we maximize the performance of the attack model, and subsequently, we defend against such an attack. This can be formulated as
\begin{align} \label{eq: adv loss}
\min_{\Phi} \max_{\Theta} \quad -L_{\mathrm{adv}} (G^\prime(\Phi),\Theta),
\end{align}
where $\Phi$ and $\Theta$ denote the parameters of the sampling component and the attack model respectively. $\Phi$ determines $G'$ by influencing $A'$.

\vpara{Defending Against a Broad Spectrum of Attacks via GHRatio. }
Eq. (\ref{eq: adv loss}) ensures that our publishing method can defend against the proposed worst-case attack. 
\YHY{However, in real-world scenarios, the released graph may face various attacks, and not all of them necessarily reach the worst case \cite{liu2024beyond, ranjan2019attacking}.}
In such scenarios, we devise a universal protection strategy via GHRatio, which serves as a measure independent of a specific attack.

Essentially, GHRatio quantifies how much information the network structure can disclose for inferring the private attribute. 
By minimizing GHRatio, we can mitigate the risk of privacy leakage in an attack-agnostic manner:

\begin{equation}
    \min_{\Phi} \quad L_{\mathrm{dis}} = \frac{1}{|V|}\sum_{v_i \in V } \Vert \mathrm{GHRatio}_i(\Phi) - \mathrm{GHRatio}^0_i \Vert,
\end{equation}
where $\mathrm{GHRatio}_i(\Phi)$ represents the new GHRatio of the sampled graph $G^\prime$. $\mathrm{GHRatio}^0_i$ represents $P(Z_i=Z_j)$, indicating the probability that nodes $v_i$ and any node $v_j (j \neq i)$ have the same private attribute. 
When $L_{\mathrm{dis}}=0$, the structural characteristic in $\mathrm{GHRatio}_i(\Phi)$ provides no benefits for attribute inference. 

Note that since $Z_i$ is given, we have $\mathrm{GHRatio}^0_i = P(Z_i)$. In practice, we propose to optimize the two prevalent cases of GHRatio, $\mathrm{GHRatio}^\text{prox}$ and $\mathrm{GHRatio}^\text{role}$, being described as:
\begin{align} \label{eq: GHR loss}
    \min_{\Phi} \quad L_{\mathrm{dis}} &=  \frac{1}{|V|} \sum_{v_i \in V } \| \mathrm{GHRatio}^\text{prox}_i (\Phi) - \hat{P}(Z_i) \| \notag \\
    & + \| \mathrm{GHRatio}^\text{role}_i(\Phi) - \hat{P}(Z_i) \|,
\end{align}
where $\hat{P}(Z_i)$ represents the empirical estimation of $P(Z_i)$.

\vpara{Regularization for Ensuring Utility.}  
To ensure the utility of the sampled graph, we aim to align the properties of $G^\prime$ with those of $G$. Thus, we incorporate a reconstruction-based regularization term to control the deviation of the sampled graph:

\begin{align} \label{eq: util loss}
\min_{\Phi} \quad L_{\mathrm{reg}}=
- \frac{1}{|E|} \sum_{(i,j) \in E} A_{i j} \log ({\mathcal{T}}_{i j}),
\end{align}
where ${\mathcal{T}}_{i j}$ denotes the sampling probability of edge $(i,j)$. By Eq. (\ref{eq: util loss}), $G^\prime$ will retain as many edges as possible from $G$.

To sum up, we formalize the overall optimization
problem:
\begin{align} \label{eq: total loss}
\min_{\Phi} \max_{\Theta} \quad
 L = - \gamma \cdot L_{\mathrm{adv}} + \eta \cdot L_{\mathrm{dis}} + \lambda \cdot L_{\mathrm{reg}},
\end{align}
where $\gamma, \eta, \lambda > 0$ are hyper-parameters.

For different scenarios, we can also modify Eq. (\ref{eq: total loss}) to obtain different variants. If the goal is to protect against the proposed worst-case attack, only retaining $L_{\mathrm{adv}}$ and $L_{\mathrm{reg}}$ would be sufficient. On the other hand, if the goal is not specifically for the worst case but to be effective against a broad range of attacks, retaining  $L_{\mathrm{dis}}$ and $L_{\mathrm{reg}}$ is suitable.

\subsection{Training Algorithm} \label{subsec: algorithm}

In the training phase, the parameters $\Phi$ of the sampling component and the parameters $\Theta$ of the proposed attack model are jointly trained. Specifically, The training algorithm iterates through the following main steps: (1) Learn $\Phi$ to minimize $-L_{\text{adv}}$, $L_{\text{reg}}$ and $L_{\text{dis}}$ while keeping $\Theta$ fixed, and (2) Learn $\Theta$ to maximize $-L_{\text{adv}}$ while keeping $\Phi$ fixed. Repeat these steps until the maximum iteration is reached. The detailed training algorithm and complexity analysis are summarized in Appendix \ref{sub: complexity}.
\vspace{0.15in}

\section{Experiments} \label{sec:exp}
In this section, we evaluate the effectiveness of both the proposed attack model and the defensive model of data publishing.

\subsection{Experiment Setting} \label{subsec: setting}
\YHY{
\vpara{Datasets.}
We conduct attribute inference and data publishing experiments on three datasets: Pokec-n, Pokec-z, and NBA \cite{ling2022learning, dai2021say}. Following prior works \cite{dai2021say}, we treat country as the private attribute in NBA, and region as the private attribute in Pokec-n and Pokec-z. Additionally, we also treat users' age as another private attribute in Pokec-n and Pokec-z, categorizing it according to the split in \cite{hu2007demographic}: Young (18-24), Young-Adult (25-34), Middle-aged (35-49), and Senior ($>$ 49). All private attributes are randomly split, with 10\% publicly available and the remaining 90\% hidden. 
In the experiments of data publishing, we also consider salary as the label in NBA and working field as the label in Pokec-n and Pokec-z \cite{ling2022learning, dai2021say}. We conduct node classification on these labels as downstream tasks (using a training-testing split of 0.1:0.9), and assess the utility of published graphs by evaluating the performance of these tasks. The statistics of the three datasets are summarized in Table \ref{tab: stats}.

\vpara{Implementation details.}
All experiments are conducted on a machine of Ubuntu 20.04 system with AMD EPYC 7763 (756GB memory) and NVIDIA RTX3090 GPU (24GB memory). All models are implemented in PyTorch version 2.0.1 with CUDA version 11.8 and Python 3.8.0.
Each experiment is repeated 5 times to report the average performance with standard deviation.

For the attack model, the encoder $\mathrm{GNN^{prox}}$ and $\mathrm{GNN^{role}}$ are implemented by two 2-layer GIN \cite{xu2018powerful} encoders, with the same model architecture. The hidden dimensions are set to 128 in Pokec-z, Pokec-n, and 64 in NBA.
The two MLPs are both implemented by 1-layer linear transformations. The model is trained by AdamW optimizer with a learning rate of 0.001 for 300 epochs in Pokec-n, Pokec-z, and 500 epochs in NBA.
For the defensive model, the sampling component consists of a two-layer Graphsage \cite{hamilton2017inductive} encoder and a two-layer MLP. 
The hidden dimension of the Graphsage encoder is 64, and the hidden dimension of the MLP is 32 in all datasets. The model is trained by AdamW optimizer with a learning rate of 0.002 for 200 epochs in Pokec-n, Pokec-z, and 100 epochs in NBA. The weight decay is consistently set as 0.0005. Both models use ReLU as the non-linear activation function.
For hyper-parameters settings. We perform a grid search of the degree similarity threshold (see Eq.~\ref{eq: structure homo}) in [0,20] with a step size of 5 in all datasets. 
We set $\lambda$
to 1, and vary  $\gamma$ and $\eta$  (see Eq.~\ref{eq: total loss})  within [10,30] in NBA and (0,20] in Pokec-z and Pokec-n,  with a step size of 5.
Our codes are available at \url{https://github.com/zjunet/GPS_KDD}.

\subsection{Experiments on Private Attribute Inference} \label{subsec: attribute inference exp}

\vpara{Baselines.}
We compare with the following attack models, which are divided into three types: 
(1) MLP: multilayer perceptions; (2) GCN \cite{kipf2016semi}, GAT \cite{velivckovic2017graph}, GraphSAGE, Mixhop \cite{abu2019mixhop} and H2GCN \cite{zhu2020beyond}: three foundational GNNs and two heterogeneous GNNs, used as comparisons to evaluate the effectiveness of the proposed attack model in capturing privacy leakage from both proximity homophily and structure-role homophily; (3) AttriInfer \cite{jia2017attriinfer}, ComInfer \cite{mislove2010you}, AI-N2V, AI-DW \cite{duddu2020quantifying}: four methods designed for private inference on graphs. AttriInfer and ComInfer are non-deep learning models, based on Markov random fields and community detection respectively. AI-N2V and AI-DW  are representation-based models that utilize Node2Vec and DeepWalk to obtain node representations.
}

\vpara{Comparison results.}
For evaluation, Table \ref{tab: inference comparison results} presents the performance results of the proposed attack model in comparison with other baseline methods on the five aforementioned private attributes. The results reveal several key insights: Firstly, the notable performance drop observed in the MLP model empirically demonstrates the crucial role of GPS. Secondly, AI-N2V and AI-DW exhibit lower performance, possibly due to their limited ability to simultaneously capture both proximity and structure-role information in these representation methods. Additionally, the performance of AttriInfer and ComInfer is weaker than that of the GNN models. The reason may be attributed to the more powerful expressive capabilities of the latter models.  Thirdly, the proposed attack model outperforms other baseline methods on all datasets, highlighting its remarkable efficacy in capturing GPS from both structure-role homophily and proximity homophily.

\begin{table}[t]
\centering
\setlength{\tabcolsep}{1.4pt}
\renewcommand{\arraystretch}{1.4}
\caption{Accuracy (age) and ROC-AUC (rest) of graph private attribute inference, where Ours denotes the proposed attack model. The best results are bolded.}
\label{tab: inference comparison results}
\resizebox{1.0\columnwidth}{!}{
\begin{tabular}{lcccccc} 
\toprule
& \multicolumn{2}{c}{Pokec-n} & \multicolumn{2}{c}{Pokec-z} & \multicolumn{1}{c}{NBA} \\
\cmidrule(lr){2-3} \cmidrule(lr){4-5} \cmidrule(lr){6-6}
& Age & Region & Age & Region & Country \\
\midrule
MLP & 54.01 (3.42) & 58.61 (1.36) & 56.54 (3.53) & 57.49 (1.10) & 51.30 (3.22) \\
GCN & 65.37 (1.03) & 81.34 (0.52) & 66.53 (1.28) & 82.88 (0.58) & 80.35 (2.21) \\
GAT & 64.20 (1.44) & 77.37 (2.28) & 65.09 (2.01) & 78.15 (2.74) & 79.03 (4.35) \\
SAGE & 65.29 (1.04) & 81.15 (0.85) & 67.14 (0.79) & 82.51 (0.69) & 80.71 (0.74) \\
MixHop & 64.46 (0.50) & 85.93 (0.87) & 66.94 (0.38) & 87.46 (0.83) & 72.90 (1.89) \\
H2GCN & 63.13 (0.27) & 80.26 (1.34) & 65.78 (1.61) & 83.80 (1.09) & 60.59 (3.31) \\
\hdashline
AttriInfer & 64.83 (0.11) & 63.85 (0.53) & 63.18 (0.39) & 63.76 (0.40) & 65.93 (2.30) \\
ComInfer & 34.18 (0.59) & 62.35 (0.29) & 39.64 (0.81) & 55.34 (0.75) & 65.97 (1.10) \\
AI-N2V & 61.13 (0.52) & 72.12 (0.42) & 59.42 (0.76) & 74.85 (0.87) & 73.00 (1.91) \\
AI-DW & 62.25 (0.69) & 75.30 (2.18) & 58.81 (0.19) & 78.90 (3.61) & 71.53 (2.88) \\
\hdashline
Ours & \textbf{66.69} (\textbf{1.03}) & \textbf{89.39} (\textbf{0.35}) & \textbf{68.80} (\textbf{0.89}) & \textbf{90.01} (\textbf{0.36}) & \textbf{83.32} (\textbf{1.36}) \\
\bottomrule 
\end{tabular}
}
\end{table}

\vpara{Ablation study.}
We conduct ablation studies to demonstrate the efficacy of each component within our attack model, including three variants: (1) Ours-prox: the inference results are solely obtained by aggregating on the entire graph; (2) Ours-role: the results are solely obtained by aggregating within each subgraph. (3) Ours-equal: dropping the GHRatios and combining the two prediction results in a 0.5:0.5 ratio.
Table \ref{tab: ablation} shows the results on the five private attributes.  Notably, the complete model consistently surpasses the performance of all variants, showing the effectiveness and necessity of simultaneously leveraging proximity homophily and structure-role homophily to capture privacy leakage.
\begin{table}[htbp]
\centering
\setlength{\tabcolsep}{1.4pt}
\renewcommand{\arraystretch}{1.4} 
\caption{Ablation study of the proposed attack model.}
\label{tab: ablation}
\resizebox{\columnwidth}{!}{ 
\begin{tabular}{lcccccc} 
\toprule
& \multicolumn{2}{c}{Pokec-n} & \multicolumn{2}{c}{Pokec-z} & NBA \\
\cmidrule(lr){2-3} \cmidrule(lr){4-5} \cmidrule(lr){6-6}
& Age & Region & Age & Region & Country \\
\midrule
Ours-prox & 65.23 (1.79) & 88.00 (0.76) & 67.53 (0.88) & 88.96 (0.42) & 82.27 (1.43) \\
Ours-role & 65.33 (2.20) & 88.37 (0.79) & 66.03 (2.48) & 89.01 (1.24) & 79.23 (2.71) \\
Ours-equal & 65.04 (2.31) & 88.76 (0.62) & 67.09 (1.62) & 89.50 (0.94) & 81.72 (1.65) \\
Ours & \textbf{66.69} (\textbf{1.03}) & \textbf{89.39} (\textbf{0.35}) & \textbf{68.80} (\textbf{0.89}) & \textbf{90.01} (\textbf{0.36}) & \textbf{83.32} (\textbf{1.36}) \\
\bottomrule
\end{tabular}
}
\end{table}

\subsection{Experiments on Data Publishing} \label{subsec: protection exp}
\YHY{
\vpara{Baselines.} We evaluate the performance of the proposed defensive model with five baselines, including: 
(1) Rand.: randomly dropping edges; 
(2) Deg./Betw. \cite{bojchevski2019adversarial}: dropping edges based on the degree or betweenness centrality in descending order; 
(3) RABV \cite{ye2020lf}:  an edge perturbation method that satisfies $\epsilon$-edge local differential privacy, where each pair of symmetric bits in the adjacency matrix is perturbed one and only one bit; 
(4) NetFense \cite{hsieh2021netfense}: a data publishing method against GNN-based inference attack on binary private attribute, the goal is to maintain data utility and protect privacy. We adopt the multi-target setting as suggested in the paper.
}

\vpara{Comparative results.}
\begin{figure*}[t]
    \centering
    {\includegraphics[width=2\columnwidth]{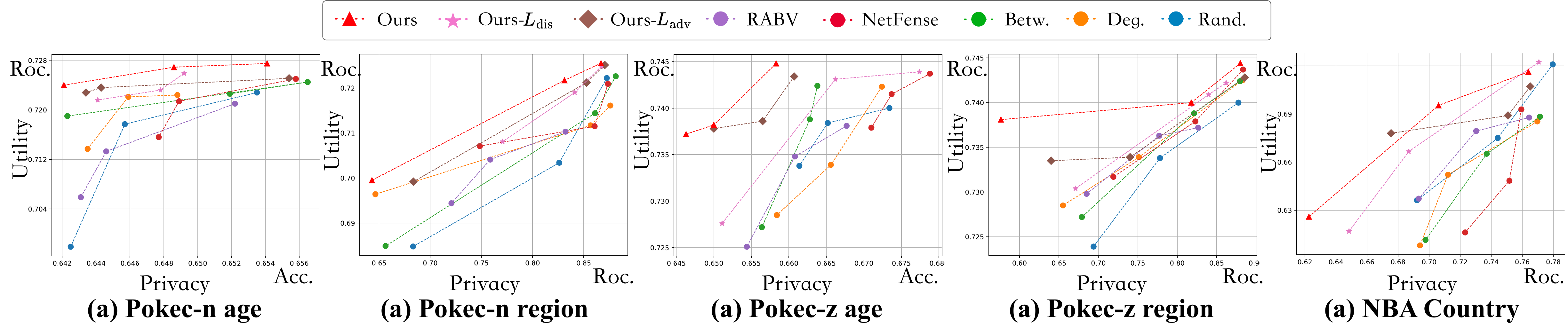}}
    \caption{Privacy-utility trade-off of our defensive model and baselines. The upper-left corner represents the ideal performance.}
    \label{fig: trade-off}
\end{figure*}
Utilizing our attack model as a worst-case adversary due to its superior performance in prior tests, we examine the privacy preservation performance by evaluating the attack models trained on the graphs generated by our defensive model and baselines. For data utility, we assess downstream task performance. Specifically, a GNN-based classifier (the same implementation as $\mathrm{GNN^{prox}}$ and MLP in \S~\ref{subsec: setting}) is employed to conduct node classification of each dataset's label on the perturbed graphs. To ensure a fair comparison, we fix the hyper-parameters for our attack model and downstream classifier, while tuning the hyper-parameters of each defensive method to explore their privacy-utility trade-offs. We select the best three trade-off points for each method and visualize them in Figure \ref{fig: trade-off}. The upper-left corner of each sub-figure represents the ideal performance, with higher downstream performance and lower attack performance. 
Note that we also report the trade-off performance of the two variants of our defensive model, namely (1) Ours-adv: only retaining $L_{\mathrm{adv}}$ and $L_{\mathrm{reg}}$ in Eq. (\ref{eq: total loss}) and (2) Ours-dis: only retaining $L_{\mathrm{dis}}$ and $L_{\mathrm{reg}}$ in Eq. (\ref{eq: total loss}).

Figure \ref{fig: trade-off} demonstrates that our defensive model consistently achieves the best privacy-utility trade-off on the five private attributes. Our variants Ours-adv and Ours-dis also demonstrate a commendable trade-off, such as in NBA country, Pokec-n age, and Pokec-z age. 
In contrast, methods such as Rand., Deg., and Eigen. do not take the private attribute into account during perturbation, thus compromising data utility. RABV exhibits suboptimal privacy preservation effects when introducing additional noise to the network structure. NetFense  fails to adequately capture privacy leakage from both structure-role homophily and proximity homophily, thereby achieving less optimal trade-offs. In addition, it presents higher computational complexity.

\vspace{0.1in}
\vpara{Evaluation of graph property change.}
To evaluate from a broader perspective, we characterize the utility by measuring the extent to which the sampled graph deviates from the original graph. Specifically, the properties of the published graph should closely resemble those of the original graph. Therefore, we employ the Maximum Mean Discrepancy (MMD) distance as our evaluation metric, comparing the degree distribution and clustering coefficient distribution of the original graph with those of the published network under our model and baselines. To ensure fairness, we tune the hyperparameters of these models to achieve comparable results in terms of privacy-preserving performance. The MMD scores in Table \ref{tab: property change} (the smaller, the better) demonstrate that our model outperforms others on each dataset except for the Pokec-z region. These results suggest that our model, while eliminating edges associated with privacy breaches, optimally preserves the remaining graph structure, effectively maintaining data utility.
\begin{table}[t]
\centering
\setlength{\tabcolsep}{1.3pt}
\renewcommand{\arraystretch}{1.4}
\caption{MMD distance of the degree distributions (d.d) and clustering coefficient distribution (c.c) between the original and the published graph of each method.}
\label{tab: property change}
\resizebox{1.0\columnwidth}{!}{
\begin{tabular}{lcccccccccc}
\toprule
& \multicolumn{4}{c}{Pokec-n} & \multicolumn{4}{c}{Pokec-z} & \multicolumn{2}{c}{NBA} \\
\cmidrule(lr){2-5} \cmidrule(lr){6-9} \cmidrule(lr){10-11}
& \multicolumn{2}{c}{Age} & \multicolumn{2}{c}{Region} & \multicolumn{2}{c}{Age} & \multicolumn{2}{c}{Region} & \multicolumn{2}{c}{Country} \\
\cmidrule(lr){2-3} \cmidrule(lr){4-5} \cmidrule(lr){6-7} \cmidrule(lr){8-9} \cmidrule(lr){10-11}
& d.d & c.c & d.d & c.c & d.d & c.c & d.d & c.c & d.d & c.c \\
\midrule
Rand. & 1.354 & 0.897 & 4.184 & 2.847 & 2.057 & 0.777 & 6.546 & 2.539 & 6.442 & 1.356 \\
Deg. & 2.834 & 0.185 & 6.077 & 1.114 & 4.190 & 0.271 & 9.196 & 0.682 & 7.769 & 1.781 \\
Betw. & 2.173 & 0.348 & 5.381 & 1.009 & 3.096 & 0.327 & 8.082 & 1.036 & 7.292 & 1.654 \\
RABV & 2.339 & 1.843 & 7.455 & 3.453 & 3.892 & 1.657 & 8.613 & 2.532 & 8.740 & 2.665 \\
NetF. & 2.093 & 0.218 & 4.138 & 1.315 & 2.229 & 0.391 & 5.765 & \textbf{0.596} & 7.673 & 1.674 \\
\hdashline
Ours & \textbf{0.192} & \textbf{0.118} & \textbf{1.759} & \textbf{0.945} & \textbf{0.392} & \textbf{0.073} & \textbf{3.279} & 0.694 & \textbf{2.498} & \textbf{0.998} \\
\bottomrule
\end{tabular}
}
\end{table}

\vpara{Evaluation of transferability.}
As our defensive model adopts the proposed inference as the worst-case adversary during training, we aim to assess its transferability. In other words, we evaluate whether the defensive model can perform effectively against other attribute inference attack models. Table \ref{tab: transferability} reports the performance of our defensive model compared with other baselines on Pokec-n age. The results show that our defensive model outperforms other methods in protecting against various attack models, demonstrating its outstanding transferability.

\YHY{Evaluation of transferability on other private attributes and more comprehensive experiments can be found in the full version.}

\begin{table}[t]
\centering
\caption{Defending against various attack models on Pokec-n.}
\label{tab: transferability}
\renewcommand{\arraystretch}{1.4}
\resizebox{1.0\columnwidth}{!}{
\begin{tabular}{lcccc}
\toprule
& AttriInfer & ComInfer & AI-N2V & AI-DW \\
\midrule
Rand. & 63.90 (0.63) & 34.01 (1.01) & 60.98 (0.39) & 61.76 (0.77) \\
Deg. & 63.20 (1.51) & 33.62 (0.52) & 60.92 (0.58) & 61.23 (0.37) \\
Betw. & 62.96 (0.63) & 33.17 (0.64) & 61.03 (0.89) & 61.42 (0.45) \\
RABV & 63.34 (0.81) & 33.49 (0.83) & 60.52 (0.45) & 61.49 (0.87) \\
NetFense & 62.84 (0.41) & 32.73 (0.59) & 60.01 (0.52) & 61.10 (0.62) \\
\hdashline
Ours & \textbf{62.02} (\textbf{0.19}) & \textbf{31.92} (\textbf{0.63}) & \textbf{59.78} (\textbf{0.11}) & \textbf{60.86} (\textbf{0.55}) \\
\bottomrule
\end{tabular}
}
\end{table}
\vspace{0.15in}

\section{Related Work} \label{sec:related}
\vpara{Private attribute inference.} Early approaches to attribute inference have primarily focused on using user-individual public attributes. These attributes include profile labels \cite{fang2015relational,li2014all}, textual content \cite{rao2010classifying,otterbacher2010inferring}, and location information from users' public posts \cite{li2014inferring}. 
These approaches heavily rely on the correlation between public and hidden private attributes to build inference models. Despite their demonstrated effectiveness, these methods often overlook valuable information from the connections between users, resulting in a noticeable performance decline. 

Subsequent explorations of attribute inference leverage network structure \cite{mislove2010you,zhang2022mli, jia2017attriinfer, duddu2020quantifying} and involve the utilization of graph propagation algorithms, such as GCN \cite{wu2020joint} and MRF \cite{jia2017attriinfer}, to facilitate the propagation of information across connected nodes. These models aggregate information from adjacent nodes \cite{zhang2022mli,jia2017attriinfer}, leverage community structures \cite{mislove2010you}, or random walk \cite{duddu2020quantifying} to infer private attributes. While they underscore the exploitation of proximity homophily \cite{al2012homophily}, they often overlook the other crucial aspect of structure-role homophily, thereby achieving less than optimal performance. 

\vspace{0.1in}
\vpara{Privacy-preserving learning on graph.}
\YHY{
In recent years, increasing attention has been given to the security and privacy issues of graph-based learning \cite{wang2024safety,liao2024value, xu2022blindfolded, zheng2021graph,xu2022unsupervised}, among which privacy-preserving techniques play a crucial role in graph data publishing. Defense methods based on anonymization \cite{ding2019novel, zhou2022structure}, sampling \cite{hsieh2021netfense}, model training \cite{wang2021privacy,li2020adversarial} and differential privacy \cite{zhu2014correlated, chen2014correlated, li2023differential, ye2020lf,daigavane2021node,sajadmanesh2023gap} have been proposed. 
The anonymization methods \cite{ding2019novel, zhou2022structure} face constraints due to the need to mitigate operational complexities and often compromise privacy and utility for efficiency.
The sampling-based method \cite{hsieh2021netfense} proposes an edge perturbation technique to defend against GNN-based inference on binary private attributes. However, it fails to consider privacy leaks from structural information and has high computational complexity.
Regarding model training-based defense methods \cite{wang2021privacy,li2020adversarial}, they often fall short when dealing with complex scenarios that require direct processing of graph data.
In addition, differential privacy (DP) is a common privacy protection technique. Early efforts \cite{yang2015bayesian, chen2014correlated, zhu2014correlated} extend DP to correlated settings, where data records are assumed to be correlated with each other (\textit{e.g.}, network structure). They primarily rely on noise injection for privacy preservation. In contrast, our method systematically addresses the attack-defense problem by considering the complex relationships and structural patterns encompassed in graph data.
Recently, DP-DGAE \cite{li2023differential} perturbs the objective function of graph auto-encoders to prevent attackers from re-identifying nodes. Local DP \cite{cormode2018privacy} allows individuals to locally perturb their graph metrics, such as node degree and adjacency list before aggregation to mitigate the risk of privacy leakage \cite{ye2020lf,daigavane2021node,sajadmanesh2023gap}. Striking a balance between utility and privacy remains a challenge for them in graph data publication.
Note that our method differs from DP in two key aspects: first, the determination of the edge sampling probability in DP is established according to predetermined mechanisms with respect to network structure. In contrast, our method learns the sampling probability based on the risks associated with GPS. Second, DP aims to preserve membership privacy, that is, altering one sample (\textit{e.g.}, node or edge) doesn't significantly change the output distribution, while the privacy we investigate in this work pertains to attribute-wise privacy.
}
\vspace{0.15in}

\section{Conclusion}
In this work, we delve into the problem of GPS and uncover the underlying mechanisms, including structure-role homophily, proximity homophily, and their intricate interplay. Based on this understanding, we introduce a novel data-centric approach for graph private attribute inference, capable of capturing privacy leaks from these mechanisms. Serving as the worst-case adversary, this method provides a comprehensive evaluation of potential privacy risks. To combat GPS, we propose a learnable graph sampling model for privacy-preserving data publishing. Our model enhances privacy security by learning the risks associated with each edge in GPS. Extensive experiments validate the effectiveness of our attack method and demonstrate the advantageous balance achieved by our defensive model between privacy preservation and utility retention.

\begin{acks}
This work was supported in part by NSFC (62206056, 92270121, 72271059, 62322606, 62441605, 72101007), SMP-IDATA Open Youth Fund, CCF-Tencent Rhino-Bird Open Research Fund, Joint Funds of Zhejiang Provincial NSFC (LHZSD24F020001), Zhejiang Province ``LingYan" Research and Development Plan Project (2024C01114), and Zhejiang Province High-Level Talents Special Support Program ``Leading Talent of Technological Innovation of Ten-Thousands Talents Program" (2022R52046).
\end{acks}
\bibliographystyle{ACM-Reference-Format}
\balance
\bibliography{ref}

\appendix
\appendix
\section{Appendix}
\subsection{Notations}
To facilitate clarity in our presentation, Table~\ref{tab: notation} summarizes major notations used in this work.
\begin{table}[h]
\renewcommand{\arraystretch}{1.1}
\centering
\caption{Major notations used in this work.}
\label{tab: notation}
\begin{tabular}{lp{5.4cm}} 
\toprule 
\textbf{Notation} & \textbf{Definition} \\
\midrule
$G,V,E$ & Graph, node set, and edge set\\ 
$X, A$& Attribute matrix and adjacency matrix \\ 
$Z_L, Z_U$ &  Known/hidden private attributes\\ 
$G^\prime, E^\prime$ & Sampled graph and sampled edge set\\
$\delta()$ & Homophily indicator \\
$\mathcal{N}(), g()$ & Neighborhood and ego network \\
$\mathrm{GNN^{prox}}, \mathrm{GNN^{role}}$& GNN encoders \\
$H^{\mathrm{prox}},H^{\mathrm{role}}$ &  Node representations\\
$S_i, \mathcal{L}_i$ & Induced subgraph and Laplacian matrix \\ 
$\Psi()$ & 1-hop graph filter \\
$\Theta,\Phi$ & Model parameters\\  
$\eta, \lambda, \gamma$ & Hyper-parameters\\ 
$\hat{Z}$ &  Inferred private attributes\\
$C$ & Number of private attribute's classes\\ 
$\mathcal{T}_{ij}, A^\prime_{ij} $ & Sampling probability, sampled adjacency \\
$P(), \hat{P}()$&  Probability and its empirical  estimation\\ 
$\sigma$ & Sigmoid function\\ 
$\oplus$ & Concatenation operation\\ 
\bottomrule 
\end{tabular}
\end{table}

\subsection{Proof of Theorem \ref{thm: local structure similarity}} \label{sub: proof}

\begin{lemma}[] \label{lemma: 1}
For any $A \in \mathbb{R}^{n \times d}$, let $a_i\in \mathbb{R}^{1 \times d}$ denote the $i$-th row of $A$, we have
\begin{equation}
\|a_i\|_2 \leq  \|A\|_2, \forall 1=1,...,n
\end{equation}

\begin{proof}
Let $e_i=[0,...,1,...,0]^T $ be a vector of zero, except for the $i$-th element. By the definition of matrix norm, we have $\|A^T e_i\|_2 \leq \|A^T\|_2 \|e_i\|_2$ and $\|A^T\|_2 = \|A\|_2 $. Then we have $\|a_i\|_2 = \|A^T e_i\|_2 \leq \|A^T\|_2 \|e_i\|_2 = \|A\|_2 $.
\end{proof}
\end{lemma}
We now give the proof of Theorem \ref{thm: local structure similarity}.
\begin{proof}  

Without loss of generality, we prove Theorem \ref{thm: local structure similarity} by instantiating the encoder as a $K$-layer GCN encoder and a 1-hop graph filter $\Psi(\mathcal{L})= Id-\mathcal{L}$. For simplicity, we denote the $l$-th layer's node representation in $S_i$ as $H^l_{S_i}$, which is obtained as 
\begin{equation}
H^l_{S_i}=\sigma(\Psi(\mathcal{L}_i)  H_{S_i}^{l-1} W^{l})
\end{equation}
where $\sigma$ denotes a $\tau_\sigma$-Lipschitz activation function, $W^{l} \in \mathbb{R}^{d \times d} $ denotes the learnable parameters in the $l$-th layer.

Assume that $S_i$ and $S_j$ have the same number of nodes, and $\max_{l} \|H^l_{S_i}\|_2 \leq \tau_h $ and $\max_{l} \|W^l\|_2 \leq \tau_w$. Then $\forall l=1,...,K$, we have
\begin{align}
\| H_{S_i}^l  - H_{S_j}^l\|_2 
\leq & \|\sigma(\Psi(\mathcal{L}_i)  H_{S_i}^{l-1} W^l)-\sigma(\Psi(\mathcal{L}_j)  H_{S_j}^{l-1} W^l) \|_2 \notag\\
\leq & \tau_\sigma \|\Psi(\mathcal{L}_i)  H_{S_i}^{l-1} W^l-\Psi(\mathcal{L}_j)  H_{S_j}^{l-1} W^l \|_2 \notag\\
\leq & \tau_\sigma \tau_w \| \Psi(\mathcal{L}_i) H_{S_i}^{l-1} -\Psi(\mathcal{L}_j) H_{S_j}^{l-1} \|_2 \notag\\
\leq & \tau_\sigma \tau_w \| \Psi(\mathcal{L}_i) H_{S_i}^{l-1} -\Psi(\mathcal{L}_j) H_{S_i}^{l-1} \notag\\
&  + \Psi(\mathcal{L}_j) H_{S_i}^{l-1} - \Psi(\mathcal{L}_j)  H_{S_j}^{l-1} \|_2 \notag\\
\leq & \tau_\sigma \tau_w \tau_h \| \Psi(\mathcal{L}_i) -\Psi(\mathcal{L}_j) \|_2 \notag \\
& + \tau_\sigma \tau_w\|\Psi(\mathcal{L}_j)\|_2 \| H_{S_i}^{l-1} -  H_{S_j}^{l-1} \|_2 
\end{align}

The above equation be equivalently rewritten as $R_l \leq a+ b R_{l-1}$, then we have
\begin{align}
    R_l &\leq a+ b R_{l-1} \notag\\
    &\leq a(b+1)+ b^2 R_{l-2} \notag\\
    & \dots \notag\\
    & \leq \frac{b^l-1}{b-1}a + b^l R_0 
\end{align}
Let $l=K$ and let $\| H_{S_i}  - H_{S_j}\|_2 = \| H_{S_i}^K  - H_{S_j}^K\|_2$ denote the representation difference in the last layer, we have
\begin{align}
\| H_{S_i}  - H_{S_j}\|_2 \leq 
& \frac{(\tau_\sigma \tau_w)^K \|\Psi(\mathcal{L}_j)\|_2^K -1}{\tau_\sigma \tau_w\|\Psi(\mathcal{L}_j)\|_2-1} \tau_\sigma \tau_w \tau_h \| \Psi(\mathcal{L}_i) -\Psi(\mathcal{L}_j) \|_2 \notag \\
& + (\tau_\sigma \tau_w)^K \|\Psi(\mathcal{L}_j)\|_2^K \| X_i  - X_j\|_2
\end{align}
Since we assume that the attribute of each node is a vector of ones, we have $\| X_i  - X_j\|_2 = 0$. Since the graph Laplacians are normalized, we assume $\min_l \| \Psi(\mathcal{L}_j) \|_2 \leq \tau_l$. Thus
\begin{align}
    \| H_{S_i}  - H_{S_j}\|_2 
    &\leq \frac{(\tau_\sigma \tau_w \tau_l)^K -1}{\tau_\sigma \tau_w \tau_l-1} \tau_\sigma \tau_w \tau_h \| \Psi(\mathcal{L}_i) -\Psi(\mathcal{L}_j) \|_2 \notag\\
    &\leq \tau \| \mathcal{L}_i -\mathcal{L}_j \|_2 
\end{align}
where $\tau = \frac{(\tau_\sigma \tau_w \tau_l)^K -1}{\tau_\sigma \tau_w \tau_l-1} \tau_\sigma \tau_w \tau_h$.

By a pooling function of mean, we obtain the final center node representation $H^\mathrm{role}_i, H^\mathrm{role}_j$ of $S_i, S_j$. From Lemma \ref{lemma: 1}, we have:
\begin{equation}
\|H^\mathrm{role}_i - H^\mathrm{role}_j \|_2 \leq \frac{1}{n}\sum_{v=1}^n \|H_{S_i,v} - H_{S_j,v} \|_2 
\leq \tau \|\mathcal{L}_i -\mathcal{L}_j \|_2
\end{equation}
which completes the proof.
\end{proof}

\subsection{Training Algorithm and Complexity Analysis} \label{sub: complexity}
\begin{algorithm}[h] 
    \caption{Learnable graph sampling method}
    \begin{algorithmic}[1]
        \REQUIRE Graph $G=(V,E,X)$, available private attributes $Z_L$, number of epochs $t$, update interval $l$, and $\gamma, \eta, \lambda$.
        \ENSURE The sampled graph $G^\prime = (V, E^\prime, X)$.
        \STATE Initialize $\text{GHRatio}^\text{prox}$ and $\text{GHRatio}^\text{role}$ for all nodes as 0.5. 
        \FOR{epoch $e=1,2,\cdots ,t$}
            \STATE Calculate  the edge sampling probability by Eq. (\ref{eq: sampling probability}). 
            \STATE Calculate the training loss by Eq. (\ref{eq: adv loss}), Eq. (\ref{eq: GHR loss}) and Eq. (\ref{eq: util loss}).
            \STATE Update the parameters $\Theta$ of the attack model by maximizing Eq. (\ref{eq: total loss}).
            \STATE Update the parameters $\Phi$ of the sampling component by minimizing Eq. (\ref{eq: total loss}).
            \IF{$e \% l == 0$}
            \STATE Update $\text{GHRatio}^\text{prox}$ and $\text{GHRatio}^\text{role}$ by Eq. (\ref{eq: feature homo}) and (\ref{eq: structure homo}).
            \ENDIF
        \ENDFOR
        \STATE Use the learned edge sampling probability to obtain $G^{\prime}$.
    \end{algorithmic}
    \label{alg: overall}
    \vspace{0.1in}
\end{algorithm}
We divide our data publishing method into four computational steps, and we provide an analysis of the time complexity for each step.

\begin{enumerate}[leftmargin = 14pt]
\item Preprocessing: In this phase, we extract the subgraphs of all nodes. Let graph $G =(V, E, X)$, the complexity of extraction is $O(n\bar{d}^k)$, where $k$ is the number of hops, $n$ is the number of nodes, $\bar{d}$ is the average degree of nodes. This step is computed only once.

\item Sampling: In this phase, we compute the sampling probability for each edge and perform graph sampling. The complexity of obtaining node representations for each node through GraphSAGE is $O(r nKt)$, where $r$ is the number of sampled neighbors,  $K$ is the number of layers, and $t$ is the number of iterations. Here, we omit the time complexity of matrix operations. Then, obtaining edge sampling probability and performing sampling has a complexity of $O(m)$, where $m$ represents the number of edges. Therefore, the total complexity of this step is $O(rnKt + m)$.

\item Inference: In this phase, we perform attribute inference on the sampled graph. Firstly, the time complexity of obtaining predictions using GIN on the entire graph is $O(nKt)$. To mitigate the time overhead caused by repeated subgraph extraction, we store the global masks of corresponding edges in each node's $k$-hop subgraph during the preprocessing step. After each sampling step, we only need to determine which edges in the $k$-hop subgraph of each node are retained based on these masks. And using GIN on the subgraph for prediction has a time complexity of $O(n\bar{d}^k Kt)$. Thus, the overall complexity of this step is $O(n\bar{d}^k Kt)$.

\item Loss calculation: the complexities of computing $L_{\mathrm{adv}}$, $L_{\mathrm{dis}}$, and $L_{\mathrm{reg}}$ are $O(n)$, $O((\bar{d}+\bar{b})n)$, and $O(m)$, Where $\bar{d}$ denotes the average degree (number of neighbors) for a node, and $\bar{b}$ denotes the average number of nodes with a similar degree to a given node. The overall complexity of this step is $O((\bar{d}+\bar{b})n+m)$
\end{enumerate}

\subsection{Datasets statistics}
We provide the statistics of the three used datasets in this work.
\begin{table} [bh]
\centering
\normalsize
\caption{The statistics of datasets}
\label{tab: stats}
\begin{tabular}{llll}
\toprule
\textbf{Dataset}& \textbf{Pokec-n} & \textbf{Pokec-z} & \textbf{NBA}  \\
\midrule
\# nodes           & 66,569  & 67,797 & 403       \\
\# node attr. & 59      & 59     & 39    \\
\# edges           & 729,129 & 882,765& 16,570    \\
Private attr.& region/age  &  region/age & country \\
Label              & working field & working field & salary\\
\bottomrule
\end{tabular}
\end{table}

\subsection{More Experiment Results}
\vpara{Case study.}
\begin{figure}[h]
    \centering
    \includegraphics[width=0.99\columnwidth]{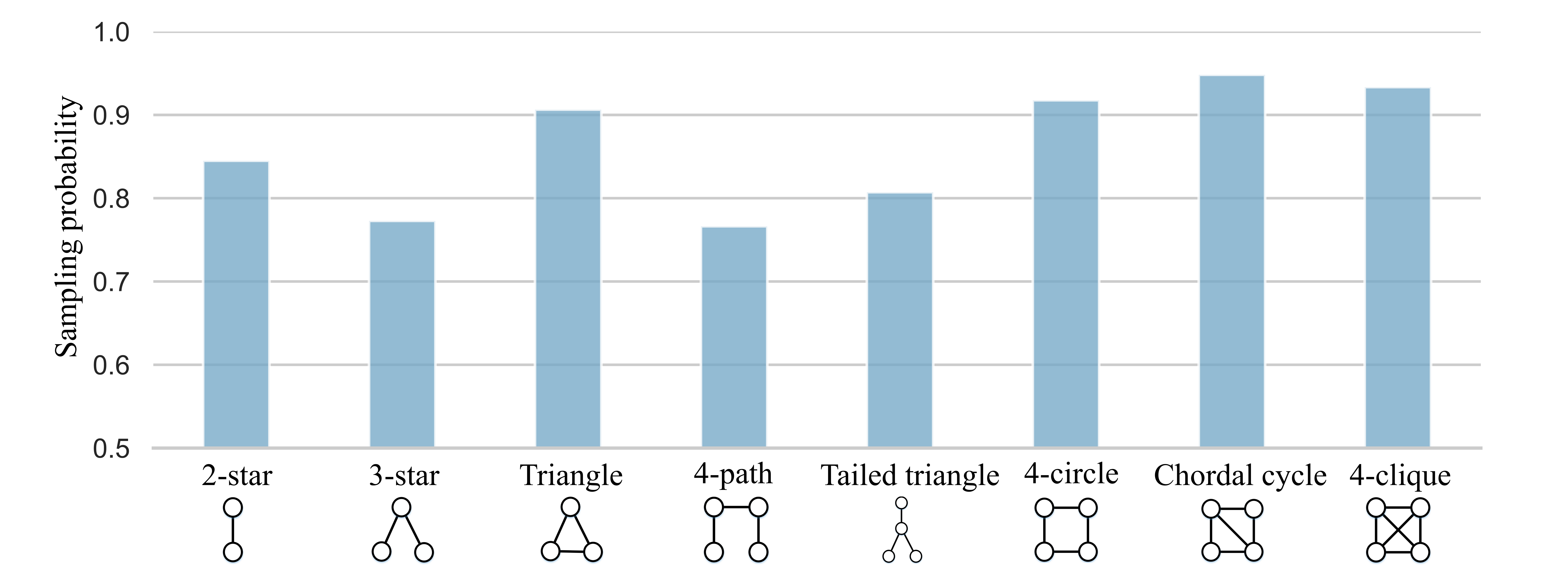}
    \caption{Average sampling probabilities among different sub-structures in Pokec-z.}
    \label{fig: case study}
\end{figure}
Taking age as the private attribute in Pokec-z, we further analyze how different local structures contribute to privacy information exposure. Based on the learned sampling probabilities of each edge, we calculate average edge sampling probabilities within eight common sub-structures (as depicted in Figure~\ref{fig: case study}). Our finding reveals a higher sampling rate for sub-structures with intricate topological characteristics, such as chordal cycles and 4-cliques. This pattern might be explained by the fact that these sub-structures contain densely connected nodes, which makes disrupting them have a pronounced impact on the entire network \cite{newman2003structure}.

\vpara{Influence of graph size and quality.}
We evaluate the influence of graph size and quality on the performance of our attack method compared to a competitive baseline, GraphSAGE. 

We first investigate the effect of graph size on the performance. We employ the Louvain community detection algorithm to partition the nodes in the NBA dataset into communities of various sizes. Specifically, we select two communities covering approximately 70\% and 40\% of the original graph. We then compare the performance of the two methods on these subsets. The results are reported in Table~\ref{tab: community_size}. We find that the attack performance decreases as the graph size shrinks, and our method consistently maintains an advantage over GraphSAGE.

\begin{table}[h]
\centering
\normalsize
\caption{Influence of graph size.}
\label{tab: community_size}
\resizebox{0.9\columnwidth}{!}{
\begin{tabular}{lccc}
\toprule
Comm. size & 100\% (Ori. graph) & $\approx$ 70\%  & $\approx$ 40\%  \\ \midrule
SAGE   & 80.71  (0.74)  & 78.56  (2.04) & 73.48 (1.53) \\
Ours & \textbf{83.32} (\textbf{1.36}) & \textbf{80.15}  (\textbf{1.65}) & \textbf{75.36}  (\textbf{1.16}) \\ \bottomrule
\end{tabular}}
\end{table}

Next, we examine the influence of graph quality by randomly masking node features to zero with different ratios. Table~\ref{tab: mask_ratio} shows the performance comparison on NBA. We observe a slight decline in our model's performance as the graph quality decreases, but it consistently outperforms GraphSAGE.

\begin{table}[h]
\centering
\normalsize
\caption{Influence of graph quality.}
\label{tab: mask_ratio}
\resizebox{0.9\columnwidth}{!}{
\begin{tabular}{lccc}
\toprule
Mask Ratio & 0\% (Ori. graph) & 20\%  & 40\%  \\ 
\midrule
SAGE       & 80.71 (0.74) & 76.84 (3.38) & 71.57 (1.95) \\
Ours       & \textbf{83.32} (\textbf{1.36}) & \textbf{81.97} (\textbf{1.35}) & \textbf{79.72} (\textbf{1.44}) \\ 
\bottomrule
\end{tabular}}
\end{table}

\vpara{Influence of GNN backbone.} 
We report the performance of our attack model with different GNNs on NBA. As shown in Table~\ref{tab: backbone}, our model maintains a consistent performance advantage over baselines, across different GNN backbones.

\begin{table}[h]
\centering
\normalsize
\caption{Performance of different GNN backbones on NBA.}
\label{tab: backbone}
\resizebox{0.9\columnwidth}{!}{
\begin{tabular}{lcccc}
\toprule
Backbone & GCN & SAGE & GAT & GIN (Ours) \\
\midrule
NBA country & 83.20  (0.83) & 82.85  (1.16) & 82.57  (0.95) & \textbf{83.32}  (\textbf{1.36}) \\
\bottomrule
\end{tabular}}
\end{table}

\vpara{Evaluation on different downstream tasks.}
We chose link prediction as another downstream task. We set the training-testing edge ratio as 0.7:0.3 and use ROC-AUC as the metric. We reported link prediction performance under similar privacy-preserving performance. Table~\ref{tab: link prediction} shows the comparison results on NBA of our model and the most competitive baseline, Netfense. The results show that our model achieves much higher link prediction performance while ensuring comparable privacy preservation.
\begin{table}[b]
\centering
\caption{Evaluation on link prediction.}
\label{tab: link prediction}
\begin{tabular}{lcc}
\toprule
Method & NetFense & Ours \\
\midrule
Link pred. $\uparrow$ & 68.37 & \textbf{76.15} \\
Privacy $\downarrow$ & \textbf{75.17} & 75.35 \\
\bottomrule
\end{tabular}
\end{table}

\vpara{Evaluation of transferability}
We provide supplementary results to highlight our model's transferability. As shown in Table \ref{tab: more transferability}, our defensive model can achieve the best protection result in most cases, and the second-best in the rest, confirming its notable transferability.
\begin{table*}[hb]
\centering
\caption{Transferability evaluation on more private attributes. The best results are bolded.}
\label{tab: more transferability}
\resizebox{1.6\columnwidth}{!}{
\begin{tabular}{llcccccc}
\toprule
Private attribute            & Adv. model & Rand.    & Deg.      & Betw.     & RABV      & NetFense  & Ours\\ \midrule
\multirow{3}{*}{Pokec-z age}   & AttriInfer & 61.67 (0.79) & 62.31 (0.61) & 62.25 (0.86) & 61.31 (0.73) & 61.46 (1.17) & \textbf{61.02 (0.49)}\\
                               & ComInfer   & 37.91 (1.24) & 36.47 (0.07) & 37.37 (0.11) & 36.42 (0.29) & 36.85 (1.59) & \textbf{36.10 (0.63)}\\
                               & AI-N2V     & 59.03 (0.23) & 58.70 (0.78) & 58.64 (0.71) & \textbf{58.24 (0.61)} & 58.91 (0.30) & 58.53 (0.48)\\
                               & AI-DW      & 57.60 (0.46) & 58.64 (0.88) & 58.70 (0.51) & 57.72 (0.32) & 57.66 (0.71) & \textbf{57.40 (0.43)}\\ \midrule
\multirow{3}{*}{Pokec-n region}& AttriInfer & 61.42 (0.48) & 60.17 (0.29) & 60.01 (0.82) & 59.52 (1.06) & 60.41 (0.46) & \textbf{58.99 (0.38)}\\
                               & ComInfer   & 61.20 (0.06) & 57.40 (0.17) & 60.91 (0.11) & 60.26 (0.07) & 59.39 (0.13) & \textbf{51.69 (0.47)}\\
                               & AI-N2V     & 69.54 (0.66) & 68.54 (0.43) & 67.92 (0.82) & 68.05 (1.98) & 68.65 (0.99) & \textbf{67.06 (0.65)}\\
                               & AI-DW      & 70.33 (1.12) & 67.79 (1.02) & 66.73 (0.46) & 68.61 (0.62) & 67.58 (0.95) & \textbf{65.29 (0.89)}\\ \midrule
\multirow{3}{*}{Pokec-z region}& AttriInfer & 61.93 (0.31) & 60.32 (0.43) & 60.86 (0.44) & 59.01 (0.48) & \textbf{58.59 (0.64)} & 58.66 (0.34)\\
                               & ComInfer   & 52.20 (0.02) & 52.40 (0.11) & 51.99 (0.28) & 51.01 (0.21) & 51.87 (0.06) & \textbf{50.81 (0.41)}\\
                               & AI-N2V     & 66.23 (0.40) & 64.42 (0.55) & 66.12 (0.56) & 64.87 (0.57) & 65.93 (0.38) & \textbf{63.92 (1.22)}\\
                               & AI-DW      & 70.01 (1.82) & 65.75 (0.90) & 69.66 (0.81) & 67.09 (0.95) & 66.44 (1.56) & \textbf{63.07 (1.06)}\\ \midrule
\multirow{3}{*}{NBA country}   & AttriInfer & 63.12 (1.12) & 58.87 (1.24) & 62.17 (1.77) & 62.42 (0.92) & 63.41 (1.29) & \textbf{57.80 (0.33)}\\
                               & ComInfer   & 63.45 (0.56) & 63.54 (1.24) & 62.81 (0.59) & 63.45 (1.01) & 63.19 (1.84) & \textbf{61.48 (1.91)}\\
                               & AI-N2V     & 69.40 (1.58) & 67.98 (0.89) & 69.44 (1.22) & 67.97 (1.37) & 68.93 (2.66) & \textbf{66.43 (1.33)}\\
                               & AI-DW      & 68.60 (2.32) & 67.89 (3.09) & 69.37 (1.06) & 68.92 (1.95) & 66.87 (2.86) & \textbf{65.41 (2.03)}\\ 
\bottomrule
\end{tabular}
}

\end{table*}

\end{document}